\documentclass[a4paper,twoside]{article}

\usepackage{epsfig}
\usepackage{subfigure}
\usepackage{calc}
\usepackage{amssymb}
\usepackage{amstext}
\usepackage{amsmath}
\usepackage{amsthm}
\usepackage{multicol}
\usepackage{pslatex}
\usepackage{apalike}
\usepackage{SCITEPRESS}
\usepackage[small]{caption}
\usepackage[hyphens]{url}
\usepackage[colorlinks=true]{hyperref}

\usepackage[table]{xcolor}
\usepackage{url}
\usepackage{pgfplots}
\usepackage{xspace}
\usepackage{booktabs}
\pgfplotsset{compat=newest}
\usepgfplotslibrary{units}

\renewcommand\vec[1]{\boldmath{#1}}
\DeclareMathOperator*{\argmin}{arg\,min} 
\usepackage{color}

\usepackage{tabularx,ragged2e}

\definecolor{building}{rgb}{0.5,0.0,0.0}
\definecolor{car}{rgb}{0.5,0.0,0.5}
\definecolor{door}{rgb}{0.5,0.5,0.0}
\definecolor{pavement}{rgb}{0.5,0.5,0.5}
\definecolor{road}{rgb}{0.5,0.25,0.0}
\definecolor{sky}{rgb}{0.0,0.5,0.5}
\definecolor{vegetation}{rgb}{0.0,0.5,0.0}
\definecolor{window}{rgb}{0.0,0.0,0.5}
\definecolor{unlabeled}{rgb}{0.0,0.0,0.0}
\newcommand\labelcolor[1]{\textcolor{white}{\cellcolor{#1}{\scriptsize #1}}}

\begin{document}

\title{Convolutional Patch Networks with Spatial Prior for Road Detection and Urban Scene Understanding}

\author{\authorname{Clemens-Alexander Brust, Sven Sickert, Marcel Simon, Erik Rodner and Joachim Denzler}
\affiliation{Computer Vision Group, Friedrich Schiller University of Jena, Jena, Germany}
\email{\{clemens-alexander.brust,sven.sickert,marcel.simon,erik.rodner,joachim.denzler\}@uni-jena.de}
{\large Project page and open source code: \url{http://cvjena.github.io/cn24/}}
}


\abstract{
  Classifying single image patches is important in many different applications, such as road detection or scene understanding. 
In this paper, we present convolutional patch networks, which are convolutional networks learned to distinguish different image patches and which can be used for pixel-wise labeling.
We also show how to incorporate spatial information of the patch as an input to the network, which allows for learning spatial priors for certain categories jointly with an appearance model.
In particular, we focus on road detection and urban scene understanding, two application areas where we are able to achieve state-of-the-art results on
the KITTI as well as on the LabelMeFacade dataset.

Furthermore, our paper offers a guideline for people working in the area and desperately wandering through all the painstaking details that render training CNs on image patches extremely difficult.

}

\onecolumn \maketitle \normalsize \vfill

\newcommand\myparagraph[1]{\paragraph{#1}}
\newcommand\classtext[1]{\textit{#1}}
\newcommand\todo[1]{\textbf{\textcolor{red}{#1}}}
\newcommand\CNN{CN\xspace}
\newcommand\CNNs{CNs\xspace}

\section{Introduction}

  In the last two years, the revival of convolutional (neural) networks (\CNN)~\cite{lecun1989cnnapplication}
  has led to a breakthrough in computer vision and visual recognition. 
  Especially the field of object recognition and detection made a huge step forward
  with respect to the final recognition performance as can be seen by the success on the large-scale image classification dataset ImageNet~\cite{krizhevsky2012imagenet}. 
  This breakthrough was possible mainly due to two reasons: (1) large-scale training data and (2) huge parallelization to speed up the learning process. 
  In general, an essential advantage of \CNNs is the automatic learning of {\it task-specific} representations of the input data, which was previously often hand-designed.

  While the majority of works focuses on applying these techniques for object classification tasks, there is another field where \CNNs can be really useful: semantic segmentation. 
  It is the task of assigning a class label to each pixel in an image.
  This is why it is also referred to as pixel-wise labeling. 
  Previous works already showed how to use \CNNs in this area, {\it e.g} for road detection~\cite{Alvarez12:RSS,Masci13:FLA}. However, the architectural choices and many critical implementation details have not been discussed and studied, although they are crucial for a high recognition performance. In our work, we therefore also give a brief list of guidelines for \CNN training and discuss several aspects important
  to get pixel-wise labeling with \CNNs running.

  \begin{figure}
    \centering
    \begin{minipage}[tb]{7.3cm}
      \centering
      \includegraphics[width=1\textwidth]{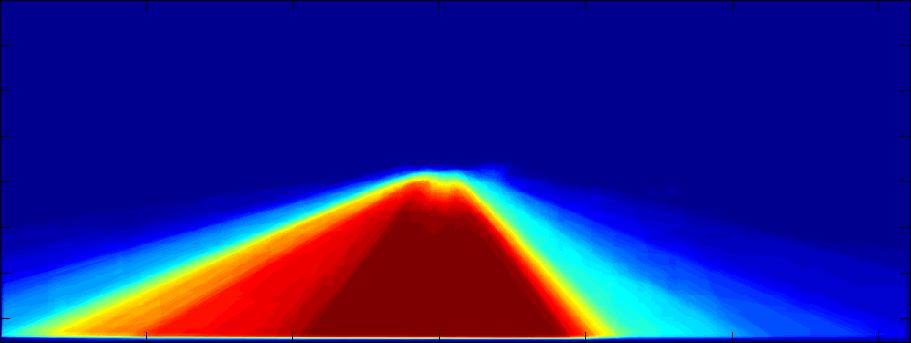}
    \end{minipage}
    \\ [1px]
    \begin{minipage}[tb]{2.385cm}
      \centering
      \includegraphics[width=1\textwidth]{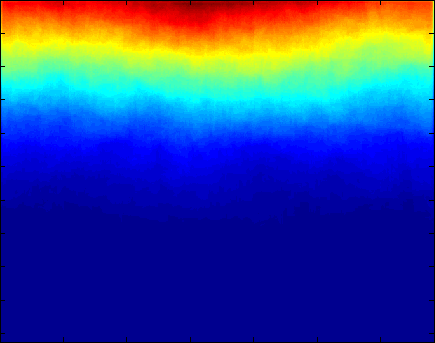}
    \end{minipage}
    \begin{minipage}[tb]{2.385cm}
      \centering
      \includegraphics[width=1\textwidth]{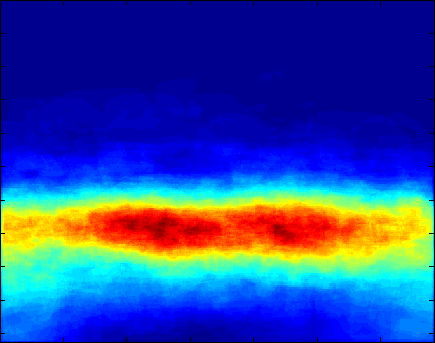}
    \end{minipage}
    \begin{minipage}[tb]{2.385cm}
      \centering
      \includegraphics[width=1\textwidth]{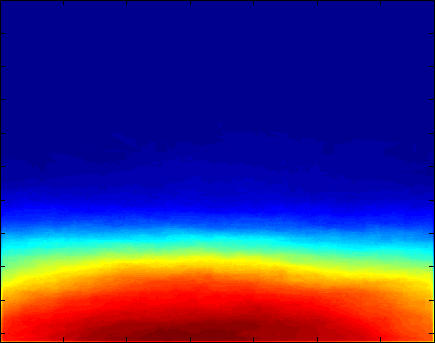}
    \end{minipage}
    \caption{Illustration of spatial bias for categories in road detection and urban scene understanding: (top) class \classtext{road} of KITTI road challenge~\cite{Geiger12:AWR}, (bottom) classes \classtext{sky}, \classtext{car} and \classtext{road} of LabelMeFacade~\cite{Froehlich12:SSM}. Warmer colors indicate higher probabilities (best viewed in color).}
    \label{fig:teaser}
  \end{figure}

  Furthermore, we show how to learn spatial priors during \CNN training, because some classes appear more frequently in some areas of an image (see Fig. \ref{fig:teaser}).
  In general, predicting the label of a single pixel requires a large receptive field to incorporate as much context information as possible.
  However, the high input dimensionality would cause a huge \CNN model with too many parameters for learning it robustly while given only 
  a small amount of training data. 
  We avoid this by incorporating absolute position information in the fully connected layers of the \CNN. 

  In this paper, we use \CNN pixel-wise labeling for the tasks of road detection and urban scene understanding.
  In road detection, the pixels of an image are classified into {\it road} and {\it non-road} parts, which is an essential task
  for autonomous driving. The challenge is the huge variability of road scenes, because of changing light conditions, surface changes, and occlusions. 
  For a qualitative and quantitative evaluation, we use the road estimation challenge of the popular KITTI Vision Benchmark Suite \cite{Geiger12:AWR}.
  Urban scene understanding goes one step further by increasing the number of categories that need to be distinguished, such as buildings, cars, sidewalks, etc. 
  We obtain state-of-the-art performance in both domains.

  In the following, we first give a brief overview of the application of \CNNs for pixel-wise labeling. Section~\ref{sec:method} introduces our \CNN models and shows how to learn spatial priors. Experiments are discussed and evaluated in Section~\ref{sec:experiments}. A summary in Section~\ref{sec:conclusions} concludes this paper.

  \vspace{-.3cm}
\section{Related work}
\label{sec:relatedwork}
  \begin{figure*}[tbp]
    \centering
    \includegraphics[width=0.9\linewidth]{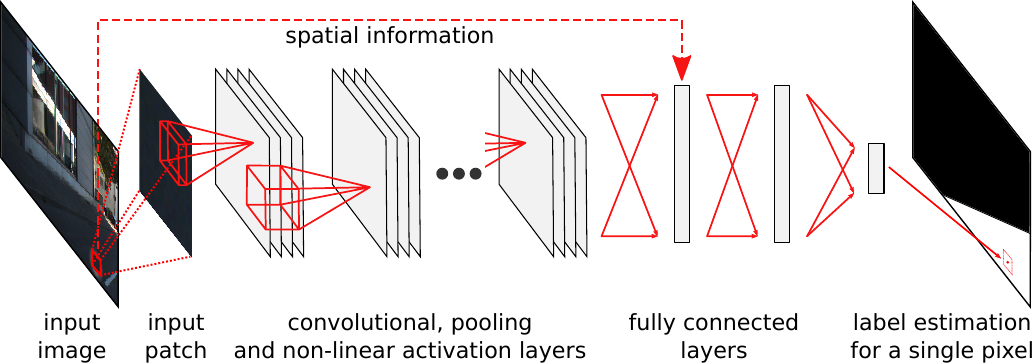}
    \caption{Example of our convolutional patch network: in addition to the visual features we also incorporate the absolute position information in the fully connected layers. The concrete architectures are given in the experimental section.} 
    \label{fig:cnn-architecture}
  \end{figure*}

  Semantic segmentation was and is an active research area with numerous publications. 
  We will present only those with relevant techniques (convolutional neural networks or randomized decision trees) or a similar scope of applications (road detection or urban scene understanding).

  \myparagraph{Semantic segmentation with \CNNs}
  The work of \cite{Couprie14:TRT} presents an approach for semantic segmentation with RGB-D images. The main idea of their work is a multi-scale \CNN comprised of multiple \CNNs for different scales
  of the images, which are all linked to the fully connected layers at the end. In contrast to their work, our approach incorporates the spatial prior information as an important cue and a possibility to learn a bias of the position of an object in the image~\cite{torralba2003contextual}.

  Instead of performing semantic segmentation by classifying image patches, \cite{gupta14rcnndepth} builds on algorithms for unsupervised region proposal generation. Each of the proposed
  regions is then classified with an SVM that makes use of features learned by \CNN using depth and geometric features as well as a \CNN trained on RGB image patches.
  Similarly, \cite{hariharan14sds} also classifies object proposals and combines a \CNN for detection and a \CNN for classifying regions.   
  In contrast to these works, we perform pixel-wise labeling and are therefore not limited to a few proposals generated by another algorithm. 

  \myparagraph{Road detection}
  Following up on their work with slow feature analysis~\cite{Kuehnl11:MRS}, the authors of \cite{Kuehnl12:SRF} propose spatial ray features to find boundaries of the road. The former work serves as a source for base classifiers which model road, boundary, and lanes. Especially the last one is very important for the method, since ray features are extracted from classifier outputs. In contrast to our work, this method strongly depends on the availability of lane markings in the scene.
  
  Another work in the field aims at the problem of changing light conditions in street scenes. In \cite{Alvarez11:RDB}, the authors compute illumination invariant images to segment the road even if the image is highly cluttered due to shadows. 
  Seeds are placed in the bottom part of the illumination invariant image where the road is supposed to be situated. All pixels that have a similar appearance to the seeds will then be classified as {\it road}. Since we learn local image filters with the \CNNs, we do not have to explicitly model illumination invariance in our approach but learn all variations from the given dataset.

  Similar to our approach, \cite{Alvarez12:RSS} applied \CNNs for the task of road detection. However, their work focuses more on the transfer of labels learned from a general image database which 
  has more images to learn from. Furthermore, they propose a texture descriptor which makes use of different color representations of the image. 
  Finally, general information acquired from road scenes and information extracted from a small area of the current image are combined in a Naive Bayes framework to classify the image.

  \myparagraph{Urban scene understanding}
  While road detection is a binary classification scenario, the task of urban scene understanding is to distinguish multiple classes like \classtext{car}, \classtext{building} and \classtext{sky}.

  In \cite{Froehlich12:SSM} so-called iterative context forests are used to classify images in a pixel-wise or region-wise manner. The method is derived from the well known random decision forests with the advantage that classification results of one level of the tree can be used in the next level as additional features. The authors of \cite{Scharwaechter13:EMS} also aim for the classification of regions. They combine appearance features of gray scale images and depth cues which are derived from dense disparity maps. They incorporate a medium-level environment model in order to obtain meaningful region hypotheses. Then, a multi-cue bag-of-features pipeline is used to classify these regions into object classes.

  There are other works that incorporate additional sources of information other than images from a single camera. In \cite{Zhang10:SSO} dense depth maps are used to compute view-independent 3D-features, {\it i.e.} {\it surface normal} and {\it height above ground}. In contrast, the authors of \cite{Kang11:MIR} make use of an additional near-infrared channel. They use hierarchical bag-of-textons in order to learn spatial context from the image. However, as these methods are closely tied to a database that provides such information, we propose a more generic approach. 

\vspace{-.4cm}
\section{Convolutional patch networks}
\label{sec:method}

  Convolutional (neural) networks (\CNNs)~\cite{lecun1989cnnapplication} belong to a family of methods, usually referred to
  as ``deep learning'' approaches, especially in the popular literature.
  The main idea is that the whole classification pipeline
  consists of one combined and jointly trained model. 
  Most recent deep learning architectures for vision are based on a single \CNN. 
  \CNNs are feed forward neural networks, which concatenate
  several layers of different types with convolutional layers playing a key role.

  \subsection{Architecture and \CNN training}
  The generic architecture of our \CNNs
  is visualized in a simplified manner in Fig.~\ref{fig:cnn-architecture}. 
  The input for our network is always a single image patch extracted around a single pixel we need to classify.
  Therefore, we use the name \textit{Convolutional Patch Network} for the method.

  The network itself is structured in multiple layers. 
  Each convolutional layer convolves the output of the previous layer with multiple learned filter masks. 
  Afterwards, the outputs are optionally combined with a maximum operation in a spatial window applied to the result of each convolution, which is known as max-pooling layer. This is followed by an element-wise non-linear activation function, such as the hyperbolic tangent or the rectified linear unit used in \cite{krizhevsky2012imagenet}.

  The last layers are fully connected layers and multiply the input with a matrix of learned parameters followed again
  by a non-linear activation function. The output of the network are scores for each of the learned categories or in the case of binary classification one score
  related to the likelihood of the positive class. 
  We do not provide a detailed explanation of the layers, since this is described in many other papers and tutorials~\cite{lecun2001}.
  In summary, we can think about a \CNN as one huge model $f(\vec{x}; \vec{\theta})$ that tries to map an image through different layers to 
  a useful output. The model is parameterized by $\theta$, which includes the weights in the fully connected layers as well as the weights of the convolution masks. 

  All parameters $\vec{\theta}$ of the \CNN are learned by minimizing the error of the network output for an example $\vec{x}_i$ compared the given ground-truth label $y_i$:
  \begin{align}
  \hat{\vec{\theta}} &= \argmin_{\vec{\theta}} \sum\limits_{i=1}^n w_i \cdot L\left( f(\vec{x}_i; \vec{\theta}), y_i \right)\enspace.
  \end{align}
  In this setting $L$ is a loss function and in our case we use the quadratic loss.

  Optimization is done with stochastic gradient descent using momentum and mini-batches of 48 training examples~\cite{krizhevsky2012imagenet}.
  The learning rate and all other hyperparameters are optimized on a validation set.

  \subsection{Incorporating spatial priors}

    As already motivated in the introduction and in Fig.~\ref{fig:teaser}, predicting the category by using only the information from a limited local receptive field 
    can be challenging and in some cases impossible. Therefore, the work of \cite{Couprie14:TRT} proposes a multi-scale \CNN approach to incorporate information
    from receptive fields with different sizes. 
    In contrast, we exploit a very common property in scene understanding. The absolute position of certain categories in the image is not uniformly distributed.
    Therefore, the position of a patch in the image can be a powerful cue. This is especially true for road detection as also validated by \cite{Fritsch13:NPM}.

    Due to this reason, we provide the normalized position of a patch as an additional input to the \CNN. In particular, the $x\in[0,1]$ and $y\in[0,1]$ coordinates are added
    as inputs to one of the fully connected layers. This can be viewed as having a smaller \CNN, which provides a feature representation of the visual information
    contained in the patch, and a standard multiple layer neural network, which uses these features in addition to the position information to perform classification.
    Whereas incorporating the position information is a common trick in semantic segmentation, with \cite{Froehlich12:SSM} being only one example, 
    combining these priors with \CNN feature learning has not been exploited before.

    \begin{figure}
{\small
\centering
\begin{tikzpicture}
\begin{axis}[width=1.0\linewidth, height=\linewidth,
xlabel={optimization epochs},ylabel={maxF performance (cross-validation)}, y unit={\%}, ymin=62.0, ymax=68.0,
legend entries={
{Tanh, heuristic initialization}, 
{Tanh, normalized initialization},
{ReLU, normalized initialization},
{ReLU, Dropout}}, legend pos=south east
]

\addplot plot [very thick]
table [
x=EPOCH, y=F1,
col sep=comma] {dltanhn.csv};

\addplot plot [very thick]
table [
x=EPOCH, y=F1,
col sep=comma] {dltanhi.csv};

\addplot plot [very thick]
table [
x=EPOCH, y=F1,
col sep=comma] {dlrelu.csv};

\addplot plot [very thick]
table [
x=EPOCH, y=F1,
col sep=comma] {dlall.csv};

\end{axis}
\end{tikzpicture}
\caption{Performance for road detection with respect to the number of optimization epochs. The performance is measured by 10 times cross-validation on the KITTI training set.}
\label{fig:deeplearning}
}
\end{figure}
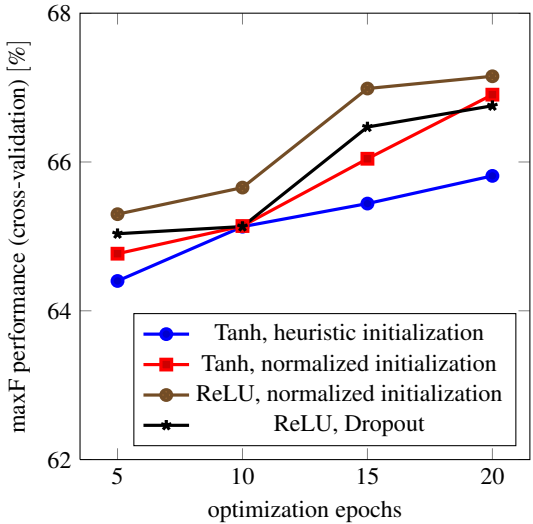

  \subsection{Software framework}

    We implemented a new open source \CNN framework specifically designed for semantic segmentation, which will be made publicly available.\footnote{This work was supported by Nvidia with a hardware donation.}
    The source code was designed from scratch in C++11 aiming at multi-core architectures and not necessarily strictly depending on GPU capabilities. 
    An important feature of the framework is the large flexibility with respect to possible \CNN architectures. 
    For example, every layer can be connected to an auxiliary input layer, which is important in our case
    to allow for the incorporation of position information or to incorporate the weight of a training example in the loss layer.

    The framework does not depend on external libraries, which makes it practical, especially for fast prototyping and 
    heterogeneous environments. However, OpenCL or fast BLAS libraries such as ATLAS, ACML, or Intel-MKL can be used to speed up 
    convolutions and other algebraic operations. Convolutions are in general realized by transforming them into matrix-vector
    products, which requires some additional memory overhead but leads to a significant speedup as also empirically validated by \cite{chellapilla2006high}. 
    For fast testing, the complete forward propagation through the network can also be accelerated by utilizing a device that computes OpenCL\footnote{\textcolor{red}{In addition to the implementation corresponding to the original VISAPP paper, our code also allows for fully convolutional network architecture.}}.

  \subsection{Important details and implementation hints}
  \label{sec:guidelines}

  Implementing our own framework allowed us to have influence on every aspect of the convolutional network in order to apply it for the task of semantic segmentation. Thereby, we made some important observations concerning parameter initialization and optimization techniques.
  
  \myparagraph{Initialization of network parameters}

    The training of networks with many layers poses a particular challenge because of the vanishing gradient issue~\cite{Glorot10:UDT}. 
    A repeated multiplication of the derivatives produces smaller and 
    smaller values. This quickly leads to numerical problems in deep networks, particularly when using single-precision floating point calculations.

    Usually, the weights in a layer with $n$ inputs are initialized randomly by sampling from a uniform distribution 
    in $[-\frac{1}{\sqrt{n}},\frac{1}{\sqrt{n}}]$, which we refer to as \textit{heuristic initialization}. 
    However, the authors of \cite{Glorot10:UDT} analyze the effect of vanishing gradients
    in detail and they derive an improved initialization scheme, known as \textit{normalized initialization}, which has an important impact on the learning
    performance. This can be seen in Figure~\ref{fig:deeplearning}, where we plot the cross-validation accuracy of our road detection application after different numbers
    of optimization epochs (an epoch are 10000 iterations with mini-batches). As can be seen, the normalized initialization leads to a better performance
    of the network after a few epochs.

  \myparagraph{Benefit of dropout and ReLU for smaller networks}
    Dropout as a regularizer is a means to prevent a network from overfitting~\cite{hinton2012improving}, which
    happens likely due to the large model complexity of the networks.
    However, the small convolutional
    net used in our approach for the task of road detection does not benefit from dropout as can be seen in Figure~\ref{fig:deeplearning}. 
    Dropout has been shown to reduce error rates significantly in larger \CNN architectures \cite{krizhevsky2012imagenet}.
    Furthermore, Figure~\ref{fig:deeplearning} also reveals that using rectified linear units~\cite{krizhevsky2012imagenet} 
    as nonlinear activations is beneficial for the task of road detection. 

  \myparagraph{Task-specific weighting of training examples}
    If a recognition approach with a high performance with respect to a task-specific performance measure is required,
    one should optimize with a learning objective that comes as close as possible to the final performance measure.
    This hint might sound simple but we give two examples in the following where this is extremely important to boost the performance.

    For the KITTI Vision road detection benchmark, performance is measured in the birds-eye view, while data is presented in ego view.
    The authors of \cite{Fritsch13:NPM} claim that the vehicle control usually happens in 2D space and therefore road detection should also be done in this space. A wrong classified pixel near the horizon in ego view represents a whole bunch of pixels in the birds-eye view. 
    To compensate for this, we need to choose weights $w_i$ for the training examples proportional to the size of the pixels after transformation in the birds-eye view.

    In urban scene understanding, we are faced with a highly imbalanced multi-class problem, since pixels labeled as building
    are more common than pixels labeled as door, for example. Therefore, performance is usually measured
    in terms of accuracy (percentage of correctly labeled pixels) and average recognition rate (average of the class-wise accuracy)~\cite{Froehlich12:SSM}.
    To focus on the average recognition rate, we weight examples according to their number of examples in the training set, \textit{i.e.} $w_i = n_{y_i}^{-1}$.

\section{Experiments}
\label{sec:experiments}

  Our experiments in semantic segmentation are evaluated on two applications: road detection and urban scene understanding, which
  are both challenging due to the high variation of possible appearances within the classes. 

  \begin{figure}
    \centering
    \includegraphics[width=0.99\linewidth]{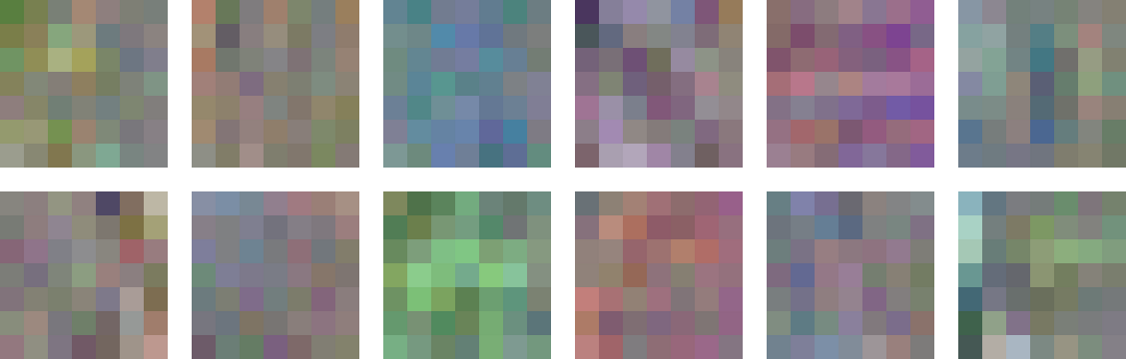}
    \caption{Convolution masks of the first layer found during learning on the KITTI road detection dataset.} 
    \label{fig:kittilayer1}
  \end{figure}
  
    \begin{figure*}[tb]
      \centering
      \includegraphics[width=0.49\linewidth]{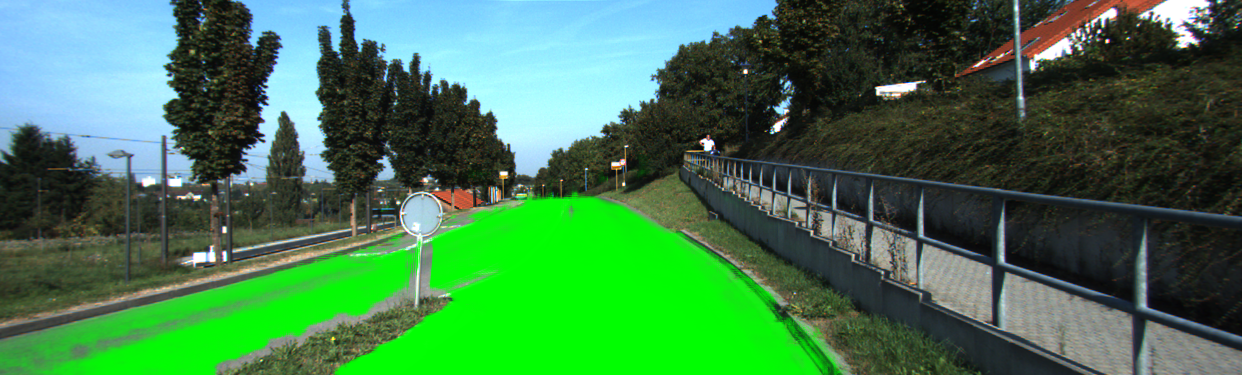}
      \includegraphics[width=0.49\linewidth]{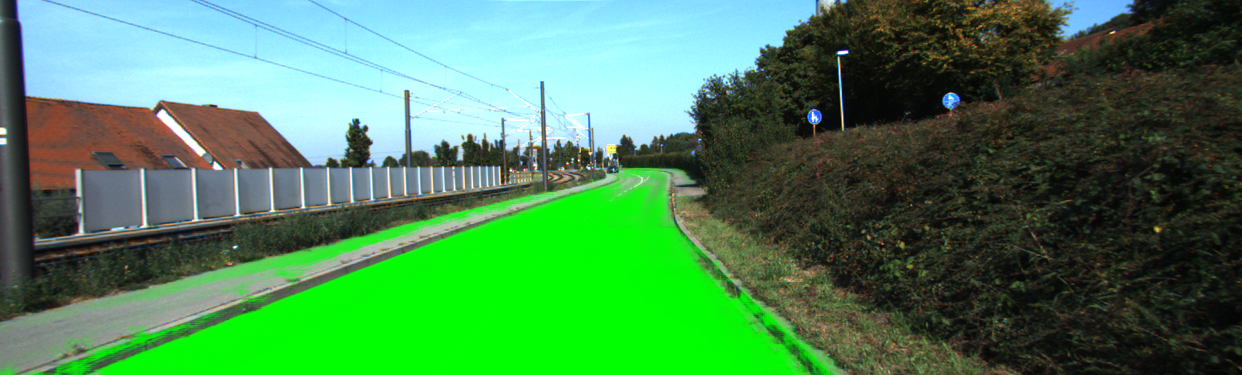}
      \caption{Qualitative results on the KITTI road detection dataset. Images show original input-data with labeled \classtext{road} pixels as green overlay. Although a spatial prior is used in our approach we are able to segment road scenes with curves or occlusions well. This figure is best viewed in color.}
      \label{fig:kitti_qual}
    \end{figure*}
 
\subsection{Road detection}
\label{sec:roaddetectionexp}

  For the task of road detection, we have to differentiate between road and non-road patches and therefore it is a typical binary classification problem.
  In recent years, the most commonly used road scene challenge is the KITTI Vision benchmark~\cite{Geiger12:AWR}. This dataset features a multi-camera setup, optical flow vectors, odometry data, object annotations, and GPS data.

  There is a specific benchmark for road detection with 600 annotated images where road and non-road parts are labeled in the image. The dataset consists of three different urban road settings: single-lane roads with markings (UM), single-lane roads without markings (UU) and multi-lane roads with markings (UMM). There are challenges for road detection and ego-lane detection.
  For this dataset, we follow the evaluation protocol given in \cite{Geiger12:AWR} and report the F1-measure as a binary classification metric which makes use of precision and recall such that both have the same weight. 
  
  \myparagraph{\CNN Architecture}
    We use the \CNN architecture listed in Table~\ref{tab:architectures} for road detection, where we classify patches of size $28 \times 28$ extracted at each pixel location. This architecture was optimized using ten-fold cross validation. An important architectural choice is the incorporation of the absolute position of the patch as an input in layer 8. This allows for learning a spatial prior of the road category. Furthermore, it is interesting to note that the first layer applies convolution masks of a rather small size of $7 \times 7$. These masks are visualized in Fig.~\ref{fig:kittilayer1} and depict certain textural and color elements that seem to be informative when distinguishing between road and non-road image patches.

      \begin{table}
        \small{
          \begin{tabular}{p{0.015\linewidth}p{0.36\linewidth}p{0.20\linewidth}p{0.2\linewidth}}
            \toprule
            {\bf\#} & {\bf Type of layer} & {\bf Road det.} & {\bf Urban sun.}\\
            \midrule
            1 & convolutional layer & $7 \times 7 \times 12$ & $7 \times 7 \times 16$\\ 
            2 & maximum pooling & $2 \times 2$ & $2 \times 2$\\
            3 & non-linear & ReLU & tanh\\
            4 & convolutional layer & $5 \times 5 \times 6$ & $5 \times 5 \times 12$\\
            5 & non-linear & ReLU & tanh\\
            6 & fully connected layer & $o=48$ & $o=64$\\
            7 & non-linear & ReLU & tanh\\
            {\bf 8} & fully connected layer & $o=192$ & $o=192$\\
                    & \textbf{+spatial prior} &&\\
            9 & non-linear & ReLU & tanh\\
            10 & fully connected layer & $o=1$ & $o=8$\\
            11 & non-linear & tanh & sigmoid\\
            \bottomrule
          \end{tabular}
        }
        \caption{\CNN architectures used for road detection (road det.) and urban scene understanding (urban sun.) along with their respective parameters. The number of outputs is denoted as $o$. The parameters for a convolutional layer are given by $w \times h \times n$, where $n$ refers to the number of spatial filters used, each of them with a size of $w \times h$. For pooling layers, the parameters determine the spatial window for which the maximum operation
        is performed.}
        \label{tab:architectures}
      \end{table}

  \myparagraph{Evaluation}

     \begin{table}
        \centering
        {\small
          \begin{tabular}{p{0.75\linewidth}c}
            \toprule
            {\bf Method} & {\bf MaxF}\\ 
            \midrule
            \CNN approach of \cite{Alvarez12:RSS}     & $73.97\%$ \\ 
            Spatial prior only \cite{Fritsch13:NPM}   & $82.53\%$ \\
            Spray features~\cite{Kuehnl12:SRF}        & $88.22\%$ \\
            \midrule
            Our approach without spatial prior                & $76.34\%$ \\
            Our approach with spatial prior                   & $86.50\%$ \\
            \bottomrule
          \end{tabular}
        }
        \caption{Results on the KITTI road detection dataset for methods that only use the camera input image for prediction.}
        \label{tab:kitti_results}
    \end{table}
 
    The quantitative results of our road detection approach are given in Table~\ref{tab:kitti_results}. We compare with the method of \cite{Kuehnl12:SRF}, which obtains the current best result on the dataset, and the \CNN method of \cite{Alvarez12:RSS}. As can be seen, we outperform the previous \CNN method by a large margin of over $10\%$. This can be mainly contributed to the spatial prior we learn with the \CNN. How important position information is for this dataset can be seen by the performance of the baseline algorithm of \cite{Kuehnl12:SRF} which uses only the pixel position during testing without any appearance features.

    As the qualitative results of our approach in Figure~\ref{fig:kitti_qual} show, we are able to segment curves although a spatial prior is incorporated. This is due to the weighting of these information which is automatically learned in the training. When allowing a very high weight for the spatial prior we would obtain similar results to the baseline algorithm. 
   
    Note that we do not make use of any additional information other than a single camera view. 
    Some competitors in the challenge incorporate data of the second camera of the stereo setup or make use of the 3D point clouds from velodyne laser scanner. Their results are not reported here but are given on the KITTI website. At the time of submission we are on place 4 of 22 in the ranking\footnote{\url{http://www.cvlibs.net/datasets/kitti/eval_road.php}, our method is named CN24}.
 
  \subsection{Urban scene understanding}
    For urban scene understanding, each pixel is classified into one of $K$ classes. Our experiments are based on the LabelMeFacade dataset~\cite{Froehlich12:SSM}, which consist of 945 images. The classes that need to be differentiated are: \classtext{building}, \classtext{window}, \classtext{sidewalk}, \classtext{car}, \classtext{road}, \classtext{vegetation} and \classtext{sky}. Furthermore, there is an additional background class named \classtext{unlabeled}, which
    we only use to exclude pixels from the training data. Since this is a multi-class classification problem, we are following \cite{Froehlich12:SSM} and use the overall recognition rate (ORR, plain accuracy) and the average recognition rate (ARR) which is the mean of class-wise accuracies.

  \myparagraph{\CNN Architecture}
      
      For urban scene understanding, we use the \CNN architecture reported in the right column of Table~\ref{tab:architectures}, which was also optimized with $50$ training examples
      and $50$ validation examples randomly selected from the LabelMeFacade dataset. 
      As in the previous experiment, we extract patches of size $28\times28$ at each pixel location.
      In contrast to the \CNN for road detection, we used the hyperbolic tangent function as a non-linearity because it requires fewer neurons to express anti-symmetric behavior.

      \begin{table}
        \centering
        {\small
          \begin{tabular}{p{0.57\linewidth}cc}
            \toprule
            {\bf Method}                            & {\bf ORR}          & {\bf ARR}\\ 
            \midrule
            RDF+SIFT \cite{Froehlich10:AFA}         & $49.06\%$          & $44.08\%$\\
            ICF~\cite{Froehlich12:SSM}              & $67.33\%$          & $56.61\%$\\
            RDF-MAP \cite{Nowozin14:IoU}            & $71.28\%$          & -\\ 
            \midrule
            {\bf Our approach}                              &                    & \\
            pure \CNN outputs                       & $67.87\%$          & $42.89\%$\\
            +spatial prior                          & $72.21\%$          & $47.74\%$\\ 
            ~+post processing                       & $74.33\%$          & $47.77\%$\\
            ~~+weighting                            & $63.41\%$          & $58.98\%$\\
            \bottomrule
          \end{tabular}
          \caption{Results on LabelMeFacade in comparison to previous work. We report overall and average recognition rates for different networks. The {\it weighted} \CNN was optimized with inverse class frequency weights.}
          \label{tab:labelmefacade_results}
        }
      \end{table}

    \begin{figure*}[tb]
      \centering
      \setlength\columnsep{0pt}
      \begin{tabular}{  >{\centering\arraybackslash}m{0.17\linewidth} 
                        >{\centering\arraybackslash}m{0.17\linewidth} 
                        >{\centering\arraybackslash}m{0.17\linewidth} 
                        >{\centering\arraybackslash}m{0.17\linewidth} 
                        >{\centering\arraybackslash}m{0.17\linewidth} 
      }
        \small{Original} & 
        \small{Ground-Truth} & 
        \small{pure \CNN outputs} & 
        \small{{\bf with} spatial prior} & 
        \small{{\bf and} post processing}
      \end{tabular}

      \includegraphics[width=0.19\textwidth]{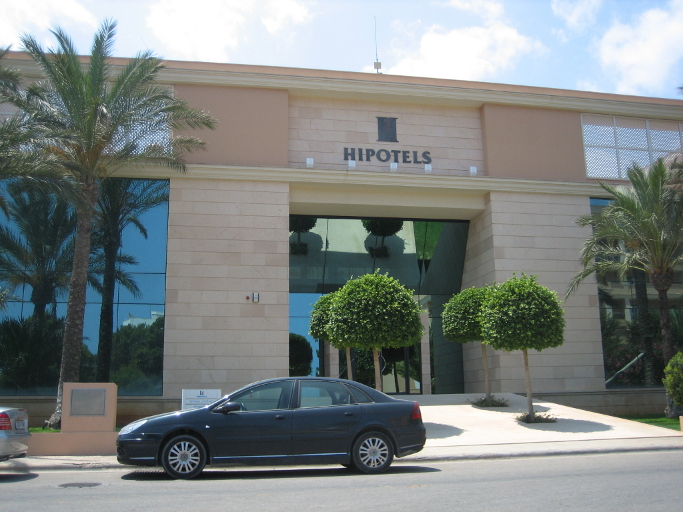}
      \includegraphics[width=0.19\textwidth]{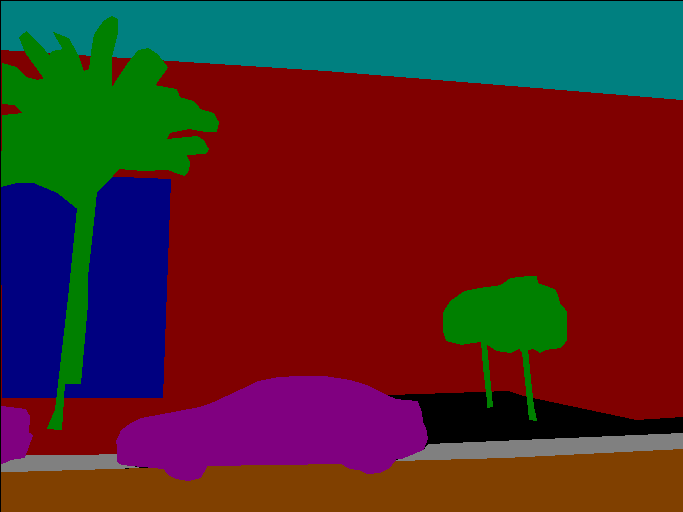}
      \includegraphics[width=0.19\textwidth]{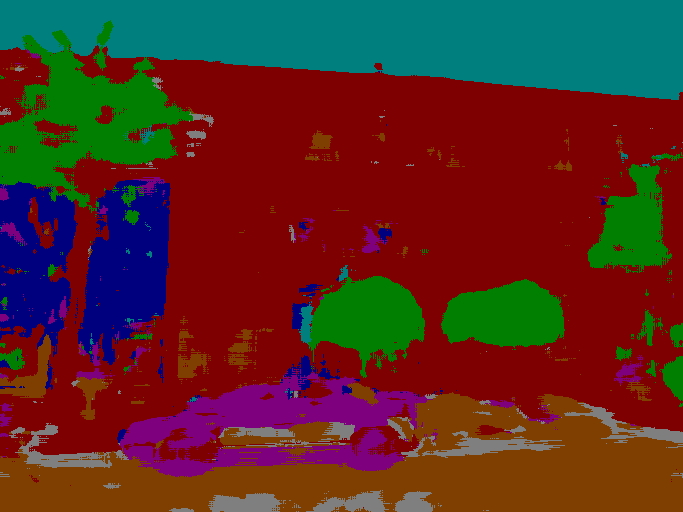}
      \includegraphics[width=0.19\textwidth]{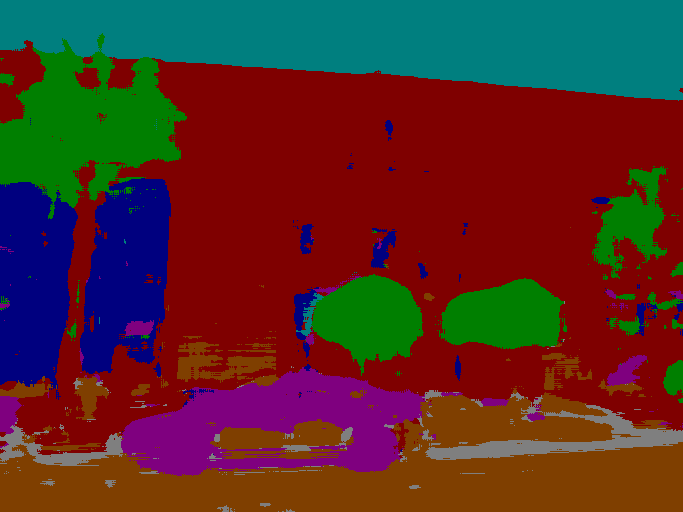}
      \includegraphics[width=0.19\textwidth]{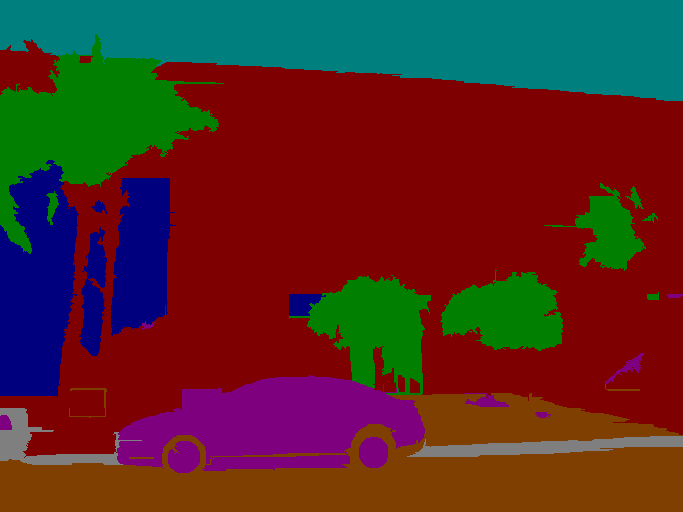}\\
      [1px]
      \includegraphics[width=0.19\textwidth]{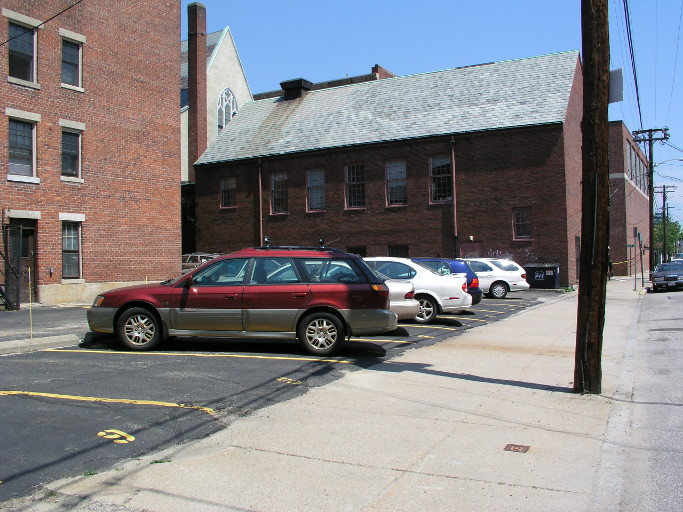}
      \includegraphics[width=0.19\textwidth]{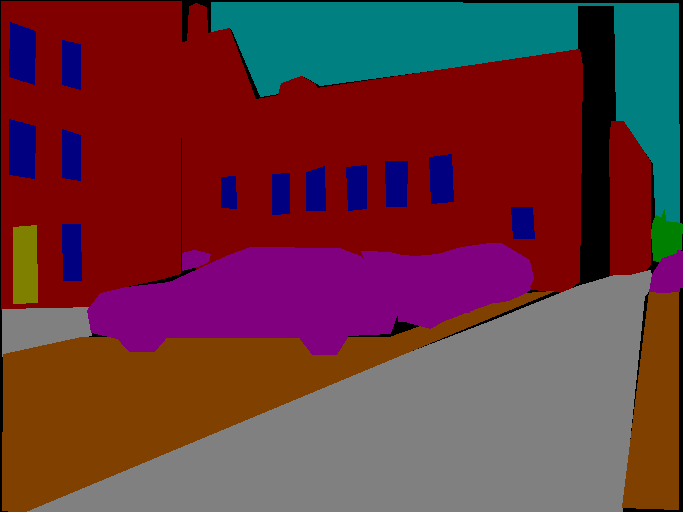}
      \includegraphics[width=0.19\textwidth]{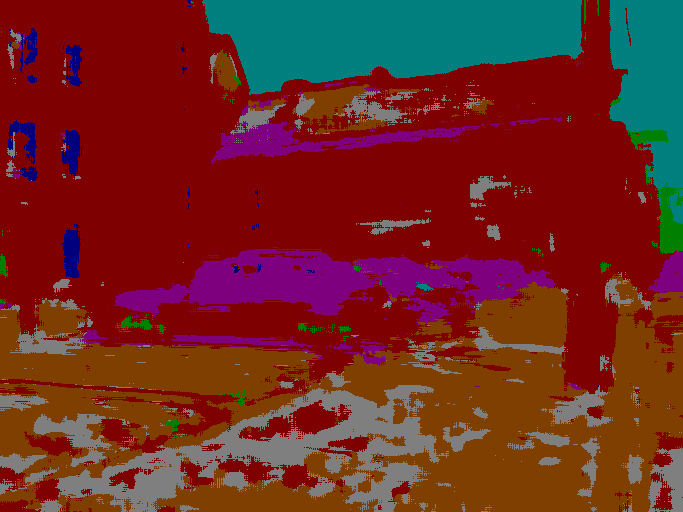}
      \includegraphics[width=0.19\textwidth]{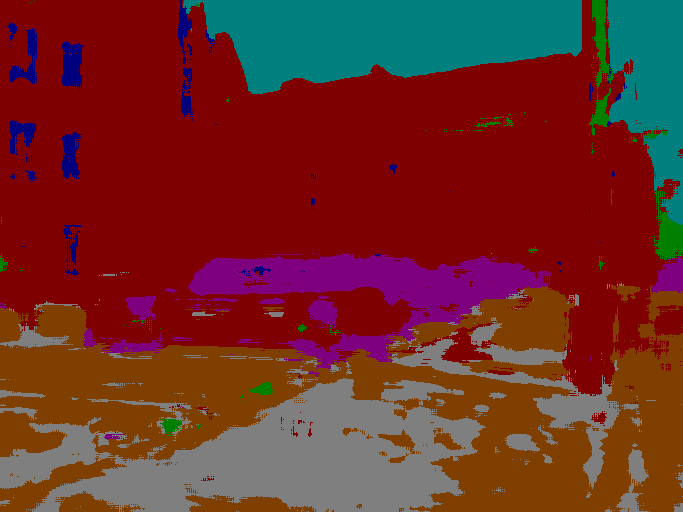}
      \includegraphics[width=0.19\textwidth]{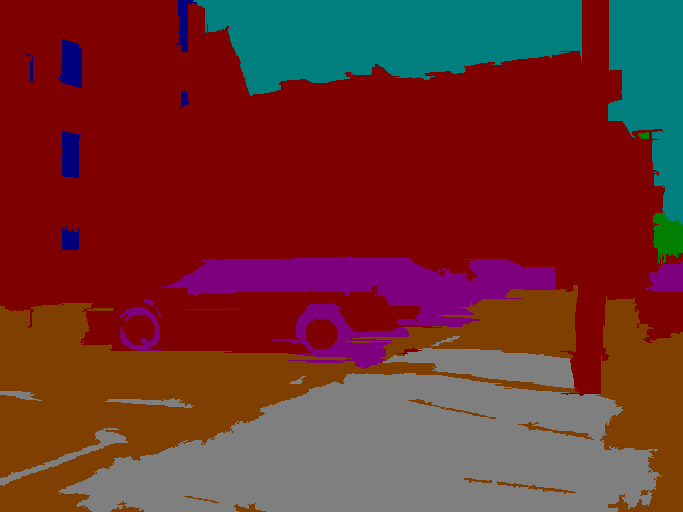}\\
      [1px]
      \includegraphics[width=0.19\textwidth]{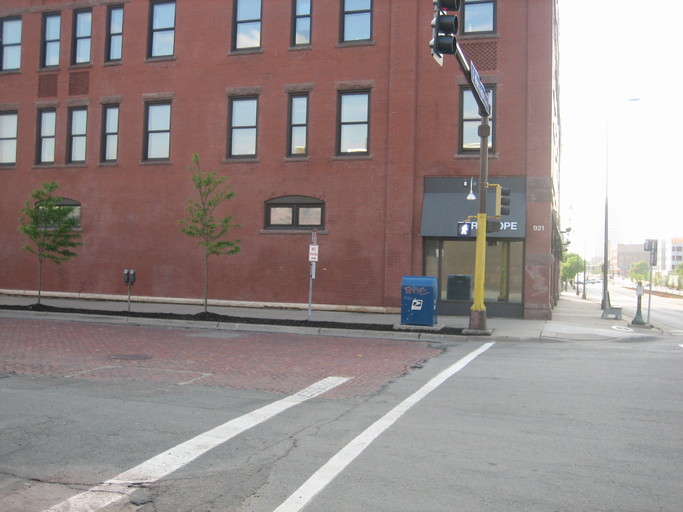}
      \includegraphics[width=0.19\textwidth]{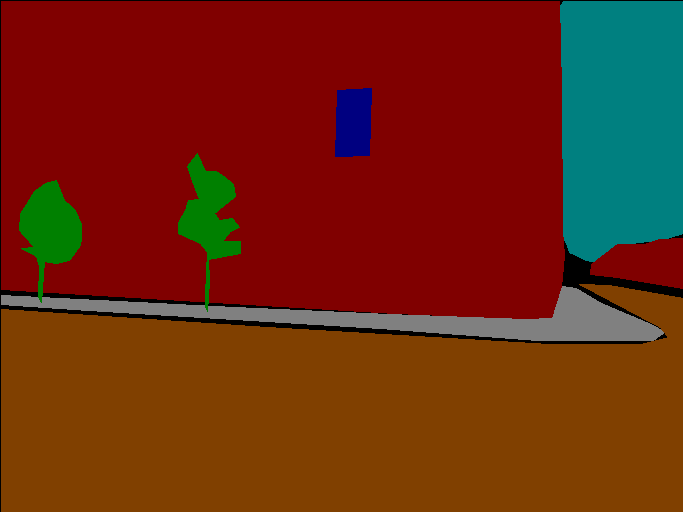}
      \includegraphics[width=0.19\textwidth]{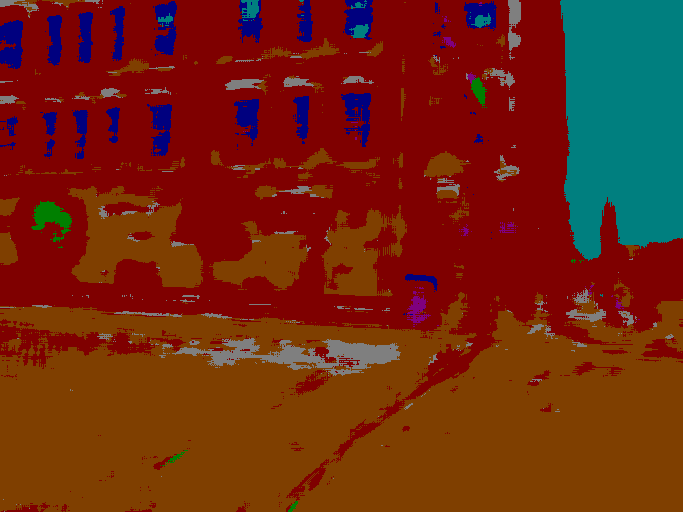}
      \includegraphics[width=0.19\textwidth]{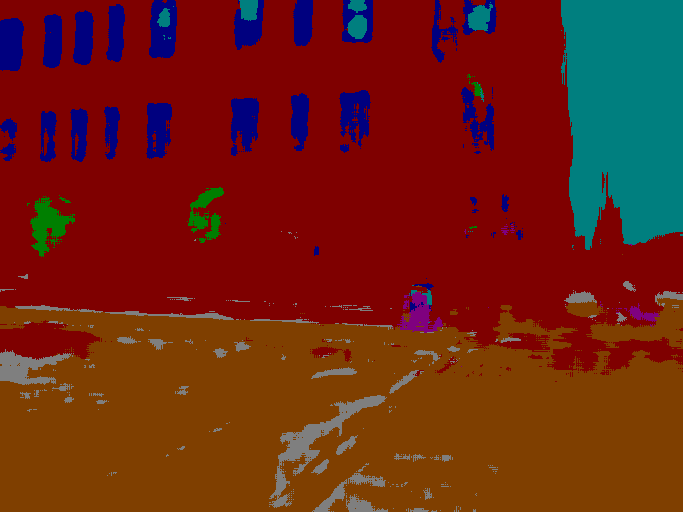}
      \includegraphics[width=0.19\textwidth]{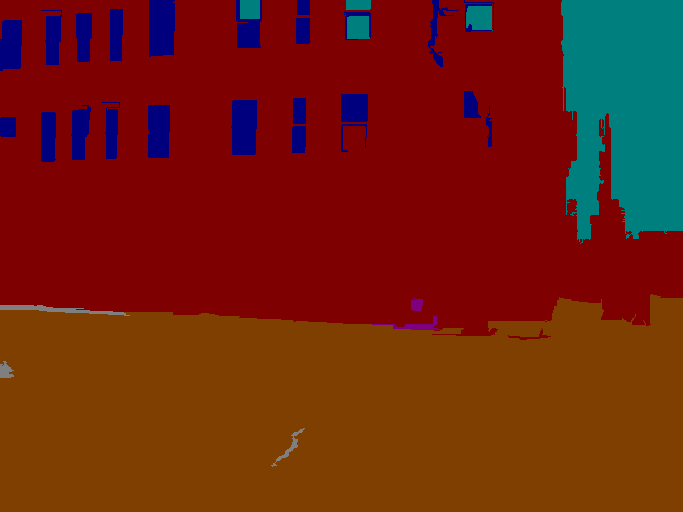}\\
      [1px]
      \includegraphics[width=0.19\textwidth]{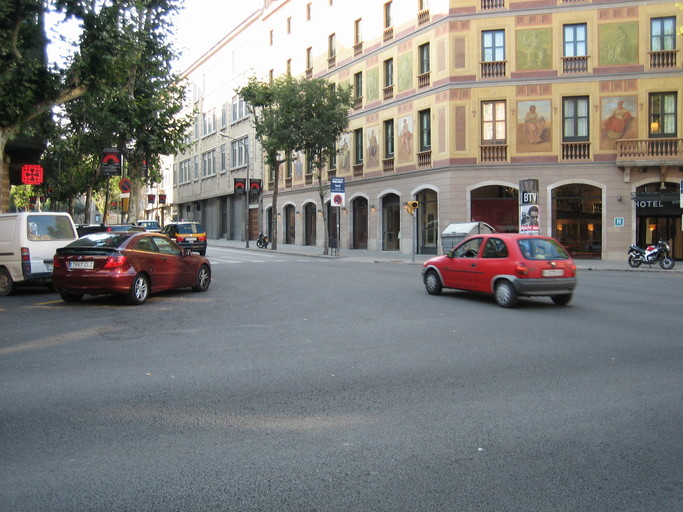}
      \includegraphics[width=0.19\textwidth]{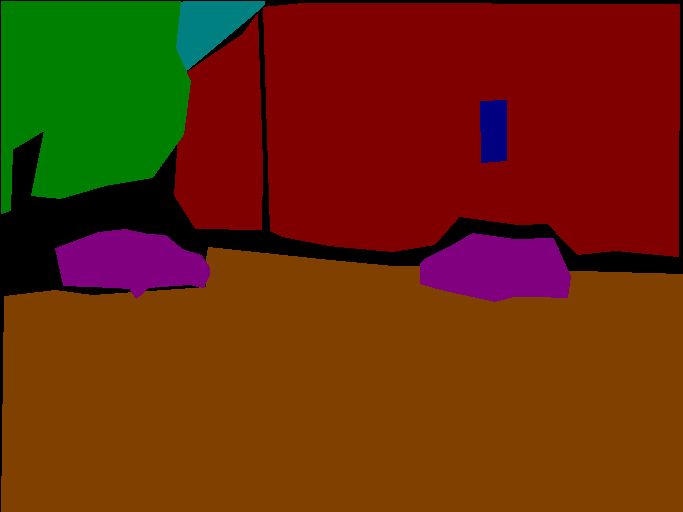}
      \includegraphics[width=0.19\textwidth]{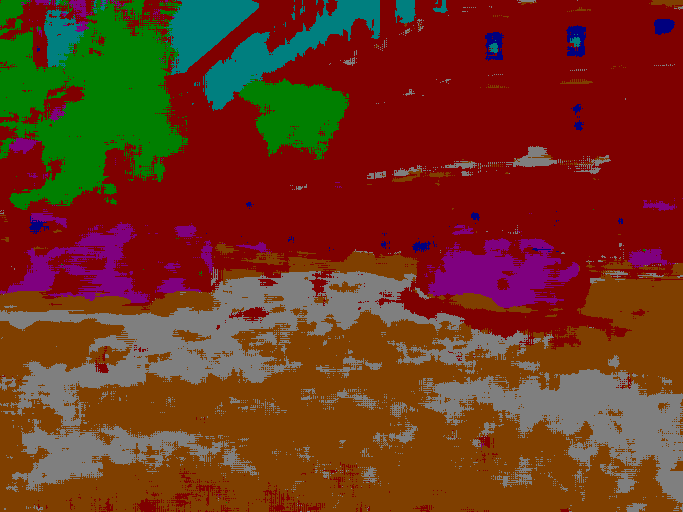}
      \includegraphics[width=0.19\textwidth]{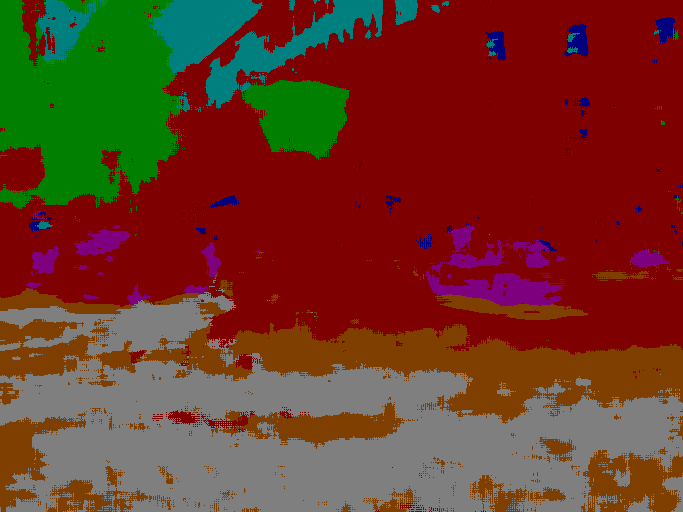}
      \includegraphics[width=0.19\textwidth]{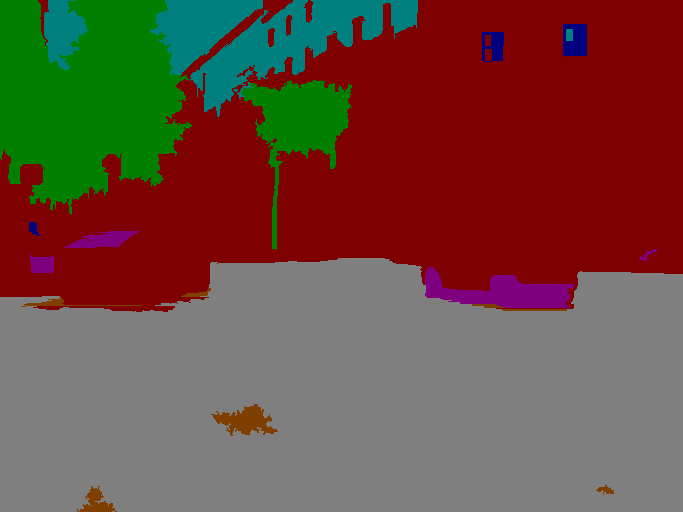}\\

      \begin{tabularx}{\linewidth}{XXXXXXXXX}
        \labelcolor{building}&
        \labelcolor{car}&
        \labelcolor{door}&
        \labelcolor{pavement}&
        \labelcolor{road}&
        \labelcolor{sky}&
        \labelcolor{vegetation}&
        \labelcolor{window}&
        \labelcolor{unlabeled}
      \end{tabularx}
      \caption{Qualitative results for the LabelMeFacade dataset. We show results of our approach with and without adding pixel positions as information for the learning procedure. As can be seen in the first three rows, these additional information improve the results (\classtext{road} vs. \classtext{building}). However, in some cases spatial priors do not help (\classtext{pavement} vs. \classtext{road}).}
      \label{tab:labelmefacade_qual}
    \end{figure*}

  \myparagraph{Evaluation}
    As can be seen in Table~\ref{tab:labelmefacade_results}, we are able to achieve state-of-the-art performance on this dataset. The spatial
    prior significantly helps again to boost the performance.
    Furthermore, we can directly see that the weighting with respect to class frequencies has a significant impact on the ARR and the ORR performance
    as already discussed in Sect.~\ref{sec:guidelines}. 

    Four segmentation examples are shown in Figure~\ref{tab:labelmefacade_qual} in comparison to the given ground-truth labels. The authors of \cite{Froehlich12:SSM}
    use a post-processing step by fusing their results with an unsupervised segmentation of the original images. The probability outputs of the classifier and the segments are combined to ensure
    a consistent label within regions.
    Since the output of our approach is scattered due to the pixel-wise labeling (column 4 and 5), we also added this post-processing step. We make use of the graph-based segmentation approach of \cite{Felzenszwalb04:EGB} with parameters $k=550$ and $\sigma=0.5$. As can be seen in the last column the results are improved with respect to object boundaries. However, this procedure can also lead to large regions with a wrong labeling (row 4). Instead of \classtext{road} the whole lower part of the image is classified as \classtext{pavement}. Both classes have a very similar appearance in most of the images.

  \vspace{-.4cm}
\section{Conclusions}
\label{sec:conclusions}

  In this paper, we showed how convolutional patch networks can be used for the task of semantic segmentation. Our approach
  performs classification of image patches at each pixel position. We analyzed different popular network architectures along with different techniques to improve the training. Furthermore, we demonstrated how spatial prior information like pixel positions can be incorporated into the learning process leading to a significant performance gain.

  For evaluation, we used two different application scenarios: road detection and urban scene understanding. 
  We were able to achieve very good results in the road detection challenge of the popular KITTI Vision Benchmark Suite. In this scenario we outperformed several competitors, even those that use stereo images or laser data. 

  For a second set of experiments, we used the dataset LabelMeFacade of \cite{Froehlich10:AFA} which is a multi-class classification task and shows very diverse urban scenes. We were again able to achieve state-of-the-art results. 
  Future work will focus on speeding up the prediction phase, since we currently need around $30s$ for each image to infer the label at each position.

  \vspace{-.5cm}

\bibliographystyle{apalike}
{\small
\bibliography{paper}}

\begin{thebibliography}{}

\bibitem[Alvarez et~al., 2012]{Alvarez12:RSS}
Alvarez, J.~M., Gevers, T., LeCun, Y., and Lopez, A.~M. (2012).
\newblock Road scene segmentation from a single image.
\newblock In {\em European Conference on Computer Vision (ECCV)}, pages
  376--389.

\bibitem[Alvarez and Lopez, 2011]{Alvarez11:RDB}
Alvarez, J.~M. and Lopez, A.~M. (2011).
\newblock Road detection based on illuminant invariance.
\newblock {\em IEEE Transactions on Intelligent Transportation Systems},
  12(1):184--193.

\bibitem[Chellapilla et~al., 2006]{chellapilla2006high}
Chellapilla, K., Puri, S., Simard, P., et~al. (2006).
\newblock High performance convolutional neural networks for document
  processing.
\newblock In {\em Tenth International Workshop on Frontiers in Handwriting
  Recognition}.

\bibitem[Couprie et~al., 2014]{Couprie14:TRT}
Couprie, C., Farabet, C., Najman, L., and LeCun, Y. (2014).
\newblock Convolutional nets and watershed cuts for real-time semantic labeling
  of rgbd videos.
\newblock {\em Journal of Machine Learning Research (JMLR)}, 15:3489--3511.

\bibitem[Felzenszwalb and Huttenlocher, 2004]{Felzenszwalb04:EGB}
Felzenszwalb, P.~F. and Huttenlocher, D.~P. (2004).
\newblock Efficient graph-based image segmentation.
\newblock {\em International Journal of Computer Vision}, 59(2):1--26.

\bibitem[Fritsch et~al., 2013]{Fritsch13:NPM}
Fritsch, J., K\"{u}hnl, T., and Geiger, A. (2013).
\newblock A new performance measure and evaluation benchmark for road detection
  algorithms.
\newblock In {\em IEEE International Conference on Intelligent Transportation
  Systems}, pages 1693--1700.

\bibitem[Fr\"{o}hlich et~al., 2010]{Froehlich10:AFA}
Fr\"{o}hlich, B., Rodner, E., and Denzler, J. (2010).
\newblock A fast approach for pixelwise labeling of facade images.
\newblock In {\em Proceedings of the International Conference on Pattern
  Recognition (ICPR)}, volume~7, pages 3029--3032.

\bibitem[Fr\"{o}hlich et~al., 2012]{Froehlich12:SSM}
Fr\"{o}hlich, B., Rodner, E., and Denzler, J. (2012).
\newblock Semantic segmentation with millions of features: Integrating multiple
  cues in a combined random forest approach.
\newblock In {\em Asian Conference on Computer Vision (ACCV)}, pages 218--231.

\bibitem[Geiger et~al., 2012]{Geiger12:AWR}
Geiger, A., Lenz, P., and Urtasun, R. (2012).
\newblock Are we ready for autonomous driving? the kitti vision benchmark
  suite.
\newblock In {\em Computer Vision and Pattern Recognition (CVPR)}, pages
  3354--3361.

\bibitem[Glorot and Bengio, 2010]{Glorot10:UDT}
Glorot, X. and Bengio, Y. (2010).
\newblock Understanding the difficulty of training deep feedforward neural
  networks.
\newblock In {\em International Conference on Artificial Intelligence and
  Statistics (AISTATS)}, pages 249--256.

\bibitem[Gupta et~al., 2014]{gupta14rcnndepth}
Gupta, S., Girshick, R., Arbel\'{a}ez, P., and Malik, J. (2014).
\newblock Learning rich features from {RGB-D} images for object detection and
  segmentation.
\newblock In {\em European Conference on Computer Vision ({ECCV})}.

\bibitem[Hariharan et~al., 2014]{hariharan14sds}
Hariharan, B., Arbel\'{a}ez, P., Girshick, R., and Malik, J. (2014).
\newblock Simultaneous detection and segmentation.
\newblock In {\em European Conference on Computer Vision ({ECCV})}.

\bibitem[Hinton et~al., 2012]{hinton2012improving}
Hinton, G.~E., Srivastava, N., Krizhevsky, A., Sutskever, I., and
  Salakhutdinov, R.~R. (2012).
\newblock Improving neural networks by preventing co-adaptation of feature
  detectors.
\newblock {\em arXiv preprint arXiv:1207.0580}.

\bibitem[Kang et~al., 2011]{Kang11:MIR}
Kang, Y., Yamaguchi, K., Naito, T., and Ninomiya, Y. (2011).
\newblock Multiband image segmentation and object recognition for understanding
  road scenes.
\newblock {\em IEEE Transactions on Intelligent Transportation Systems},
  12(4):1423--1433.

\bibitem[Krizhevsky et~al., 2012]{krizhevsky2012imagenet}
Krizhevsky, A., Sutskever, I., and Hinton, G.~E. (2012).
\newblock Imagenet classification with deep convolutional neural networks.
\newblock In {\em Advances in neural information processing systems (NIPS)},
  pages 1097--1105.

\bibitem[K\"{u}hnl et~al., 2011]{Kuehnl11:MRS}
K\"{u}hnl, T., Kummert, F., and Fritsch, J. (2011).
\newblock Monocular road segmentation using slow feature analysis.
\newblock In {\em Proceedings of the IEEE Intelligent Vehicles Symposium},
  pages 800--806.

\bibitem[K\"{u}hnl et~al., 2012]{Kuehnl12:SRF}
K\"{u}hnl, T., Kummert, F., and Fritsch, J. (2012).
\newblock Spatial ray features for real-time ego-lane extraction.
\newblock In {\em IEEE Conference on Intelligent Transportation Systems}, pages
  288--293.

\bibitem[LeCun et~al., 1989]{lecun1989cnnapplication}
LeCun, Y., Boser, B., Denker, J.~S., Henderson, D., Howard, R.~E., Hubbard, W.,
  and Jackel, L.~D. (1989).
\newblock Backpropagation applied to handwritten zip code recognition.
\newblock {\em Neural computation}, 1(4):541--551.

\bibitem[LeCun et~al., 2001]{lecun2001}
LeCun, Y., Bottou, L., Bengio, Y., and Haffner, P. (2001).
\newblock Gradient-based learning applied to document recognition.
\newblock In {\em Intelligent Signal Processing}, pages 306--351. IEEE Press.

\bibitem[Masci et~al., 2013]{Masci13:FLA}
Masci, J., Giusti, A., Ciresan, D.~C., Fricout, G., and Schmidhuber, J. (2013).
\newblock A fast learning algorithm for image segmentation with max-pooling
  convolutional networks.
\newblock {\em arXiv preprint arXiv:1302.1690}.

\bibitem[Nowozin, 2014]{Nowozin14:IoU}
Nowozin, S. (2014).
\newblock Optimal decisions from probabilistic models: the
  intersection-over-union case.
\newblock In {\em Computer Vision and Pattern Recognition (CVPR)}.

\bibitem[Scharwaechter et~al., 2013]{Scharwaechter13:EMS}
Scharwaechter, T., Enzweiler, M., Franke, U., and Roth, S. (2013).
\newblock Efficient multi-cue scene segmentation.
\newblock In {\em German Conference on Pattern Recognition (GCPR)}, Lecture
  Notes in Computer Science, pages 435--445.

\bibitem[Torralba, 2003]{torralba2003contextual}
Torralba, A. (2003).
\newblock Contextual priming for object detection.
\newblock {\em International Journal of Computer Vision (IJCV)},
  53(2):169--191.

\bibitem[Zhang et~al., 2010]{Zhang10:SSO}
Zhang, C., Wang, L., and Yang, R. (2010).
\newblock Semantic segmentation of urban scenes using dense depth maps.
\newblock In Daniilidis, K., Maragos, P., and Paragios, N., editors, {\em
  European Conference on Computer Vision (ECCV)}, pages 708--721.

\end{thebibliography}

\end{document}